\documentclass[11pt]{article}
\usepackage{coling2020}
\usepackage{times}
\usepackage{url}
\usepackage{latexsym}

\usepackage{microtype}
\usepackage[utf8]{inputenc} % UTF8 encoding
\usepackage{graphicx}
\usepackage{tabularx}
\usepackage{soul}
\usepackage{cuted}
\usepackage{hyperref}
\usepackage{xstring}
\usepackage{todonotes}
\usepackage{makecell}
\usepackage{xspace}
\usepackage{amsmath}
\usepackage{multicol}
\usepackage{multirow}

\colingfinalcopy % Uncomment this line for the final submission

\newcommand{\fmri}{fMRI\xspace}
\newcommand{\KB}{KBs\xspace}

\title{Comparative Probing of Lexical Semantics Theories\\for Cognitive Plausibility and Technological Usefulness}

\author{\\
  \textit{University of Lisbon}\\
            NLX Group\\ 
            Faculdade de Ciências\\
            Lisbon, Portugal\\%[0.5ex]
            
    \\\And
    António Branco, João Rodrigues, Małgorzata Salawa,$^{+2}$ Ruben Branco, Chakaveh Saedi,$^{+3}$\\
    $^2${\it AGH University of Science}\\ {\it and Technology}\\
    Faculty of Computer Science%, \\Electronics and Telecommunications, 
    \\Kraków, Poland\\%[0.5ex]
    {\tt antonio.branco@di.fc.ul.pt}
    \\\And
    \\
    $^3${\it Macquarie University}\\
            Department of Computing\\
            Sydney, Australia
    }
  
\date{}
\begin{document}
\maketitle
\begin{abstract}

  Lexical semantics theories differ in advocating that the meaning of words is represented as an inference graph, a feature mapping or a vector space, thus raising the question:  is it the case that one of these approaches is superior to the others in representing lexical semantics appropriately? Or in its non antagonistic counterpart: could there be a unified account of lexical semantics where these approaches seamlessly emerge as (partial) renderings of (different) aspects of a core semantic knowledge base?
\\
In this paper, we contribute to these research questions with a number of experiments that systematically probe different lexical semantics theories for their levels of cognitive plausibility and of technological usefulness.
\\
The empirical findings obtained from these experiments advance our insight on lexical semantics as the feature-based approach emerges as superior to the other ones, and arguably also move us closer to finding answers to the research questions above.
\end{abstract}

\section{Introduction}
Lexical semantics is at the core of language science and technology as the meaning of expressions results from the meaning of their lexical units and the way these are combined. How to represent the meaning of words is a central topic of inquiry 
and three broad families advocate that lexical semantics is represented as a semantic network \cite{quillan1966semantic}, a feature-based mapping \cite{minsky1975framework,bobrow1975}, or a semantic space \cite{harris1954distributional,osgood1957measurement,Miller1991ContextualCO}.

In a semantic network approach, the meaning of a lexical unit is represented as a node in a graph whose edges categorically encode different types of semantic relations holding among the units (e.g. hypernymy, meronymy, etc.). 
In a feature-based model, the semantics of a lexicon is represented by a map where a key is the lexical unit of interest and the respective value is a set of other units denoting characteristics prototypically associated with the denotation of the unit in the key (e.g. color, usage, etc.). 
Under a semantic space perspective, the meaning of a lexical unit is represented by a vector in a high-dimensional space (aka word embedding), whose components are ultimately based on the frequency of co-occurrence with other units, i.e. on their linguistic contexts of usage.

The motivation for these theories is to be found in their different suitability and success in explaining a range of empirical phenomena in terms of how these are manifest in ordinary language usage and also how they are elicited in laboratory experimentation. 
These phenomena are related to the acquisition, storage and retrieval of lexical knowledge (e.g. the spread activation effect \cite{meyer1971facilitation}, the fan effect \cite{anderson1974retrieval} among many others) and to the interaction with other cognitive faculties or tasks, as categorization \cite{estes1994classification}, reasoning \cite{rips1975inductive}, problem solving \cite{holyoak1987surface}, learning \cite{ross1984remindings}, etc.

These different approaches have inspired a number of initiatives to build repositories of lexical knowledge.
Popular representatives are, for semantic networks, WordNet \cite{fellbaum1998wordnet}, for feature-based models, SWOW \cite{de2019small}, and for semantic spaces, word2vec \cite{mikolov2013distributed} a.o. 
Different knowledge bases (KBs) are rooted in different empirical sources: WordNet is based on lexical intuitions of human experts; the information in SWOW is evoked from laypersons cued with lexical units; and word2vec reflects the co-occurrence frequency of words in texts.

Against this background, a fundamental research question is: could there be a unified account of lexical semantics such that the above approaches seamlessly emerge as (partial) renderings of (different) aspects of the same core semantic knowledge base? Or in its
antagonistic counterpart: is it the case that one of the above approaches is superior to the others in representing lexical semantics appropriately?

In this paper we contribute to these research questions with experiments consisting of two phases. First, different lexical semantic \KB, instantiating different lexical semantic theories, are converted to a common representation. The map-based dataset (SWOW) is converted to a graph-based representation and the two graph-represented datasets (SWOW and WordNet) are then converted to a vector-based representation.
Second, to assess the appropriateness of these knowledge bases, the respective word embeddings are evaluated by means of the performance in language processing tasks where they are integrated and where lexical knowledge plays a crucial role.

\begin{figure*}[t]
%\vspace{-0.5cm}
\centering
\includegraphics[width=.47\textwidth]{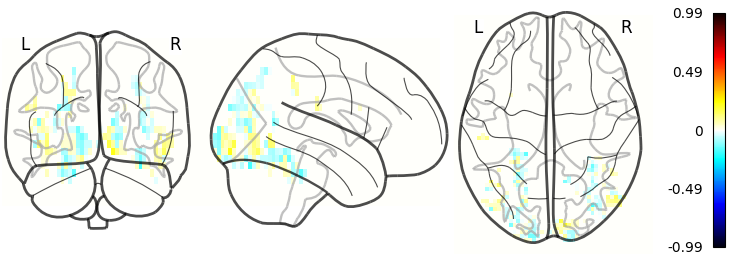}
\includegraphics[width=.47\textwidth]{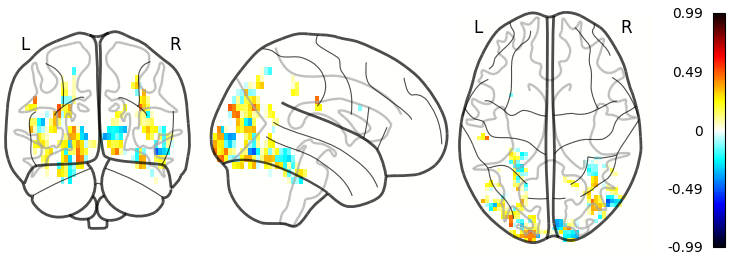}
\vspace{-0.4cm}
\caption{fMRI patterns in Participant 1 for word \textit{eye}: \textbf{predicted} (left) with WordNet 60k embeddings via matrix factorization, cf. \S  \ref{subsect:factorisation} below; \textbf{observed} (right).}
\label{fig:fmri}
\end{figure*}

We resort to the task of predicting brain activation patterns from semantic representations of words, illustrated in Figure \ref{fig:fmri}, 
to assess the lexical \KB. If a KB k1 is more successful than another k2 in supporting these tasks, this indicates that k1 has higher cognitive plausibility, likely being superior at encoding the meaning of words.

Second, we resort also to the task of determining the semantic similarity between words from their formal semantic representations. 
Though it may arguably be less well grounded on actual cognitive representation of lexical meaning given the empirical frailty of the similarity scores \cite{faruqui-etal-2016-problems}, this has been a popular task for the intrinsic evaluation of word embeddings.

Third, for extrinsic evaluation, we resort to downstream Natural Language Processing (NLP) tasks.

As reported in the present paper, the findings from these experiments indicate that the feature-based approach emerges as superior to the other approaches to the representation of lexical meaning.

\section{Related Work}

%
% The following footnote without marker is needed for the camera-ready
% version of the paper.
% Comment out the instructions (first text) and uncomment the 8 lines
% under "final paper" for your variant of English.
% 
\blfootnote{
    %
    % for review submission
    %
    \hspace{-0.65cm}  % space normally used by the marker
    This work is licensed under a Creative Commons Attribution 4.0 International Licence,
    %
    % % final paper: en-uk version 
    %
    % \hspace{-0.65cm}  % space normally used by the marker
    % This work is licensed under a Creative Commons 
    % Attribution 4.0 International Licence.
    % Licence details:
    % \url{http://creativecommons.org/licenses/by/4.0/}.
    % 
    % % final paper: en-us version 
    %
    % \hspace{-0.65cm}  % space normally used by the marker
    % This work is licensed under a Creative Commons 
    % Attribution 4.0 International License.
    % License details:
    % \url{http://creativecommons.org/licenses/by/4.0/}.
}

In \cite{mitchell2008predicting}, the meaning of each word $w$ was represented by semantic features given by the normalized co-occurrence counts in a big corpus of $w$ with a set of 25 verbs related to basic sensory and motor activities.
For each word, the respective \fmri activation level at every voxel is calculated as a weighted sum of each of the 25 semantic features, where the weights are learned by regression to maximum likelihood estimates given the observed \fmri data. Mean accuracy of 0.77 was reported for the 60 words and 9 subjects in the task of predicting brain activation originated by the exposure to words. 

In an initial period, different authors focused on different ways to set up the features supporting this task. Jelodar et al.~\shortcite{jelodar2010wordnet} used the same 25 features but resorted to relatedness measures based on WordNet.
As features, Fernandino et al. \shortcite{fernandino2015predicting} used instead 5 sensory-motor experience-based attributes, and the relatedness scores between the stimulus word and the attributes were based on human ratings.
Binder et al. \shortcite{binder2016toward}, in turn, used 65 attributes with relatedness scores crowdsourced from over 1,700 participants.

As embeddings became popular, authors moved from features to word embeddings.
Murphy et al.~\shortcite{murphy2012selecting} found their best results with dependency-based embeddings. Anderson et al. \shortcite{anderson2017visually} used word2vec together with a visual model built with a CNN on the Google Images dataset.

Recently, Abnar et al. \shortcite{abnar2017experiential} evaluated 8 embeddings in predicting fMRI patterns:
the experiential embeddings of Binder et al. \shortcite{binder2016toward};
the non-distributional feature-based embeddings of Faruqui et al. \shortcite{faruqui2015non};
and 5 different distributional embeddings, namely word2vec \cite{mikolov2013efficient}, fastText \cite{bojanowski2016enriching}, dependency-based word2vec \cite{levy2014dependency}, GloVe \cite{pennington2014glove} and LexVec \cite{salle2016matrix}; as well as the vectors from \cite{mitchell2008predicting}.
The dependency word2vec achieved the best performance among the embeddings, while the Mitchell's et al. \shortcite{mitchell2008predicting} seminal approach with 25 features ``is doing slightly better on average'' than the other approaches.

In contrast to these papers, the goal here is not to beat the state of the art in brain activation prediction but to probe lexical semantic theories with the help of this task.

\section{Common representations}
\label{sect:commonRepres}

To proceed with such comparative probing, a first step consists in the conversion of the different lexical \KB into the common representation format of semantic vector spaces, which we describe in this section. 

\subsection{From lexical maps to graphs}

\begin{figure*}[h]
%\vspace{-1cm}
\centering
\includegraphics[width=.36\textwidth]{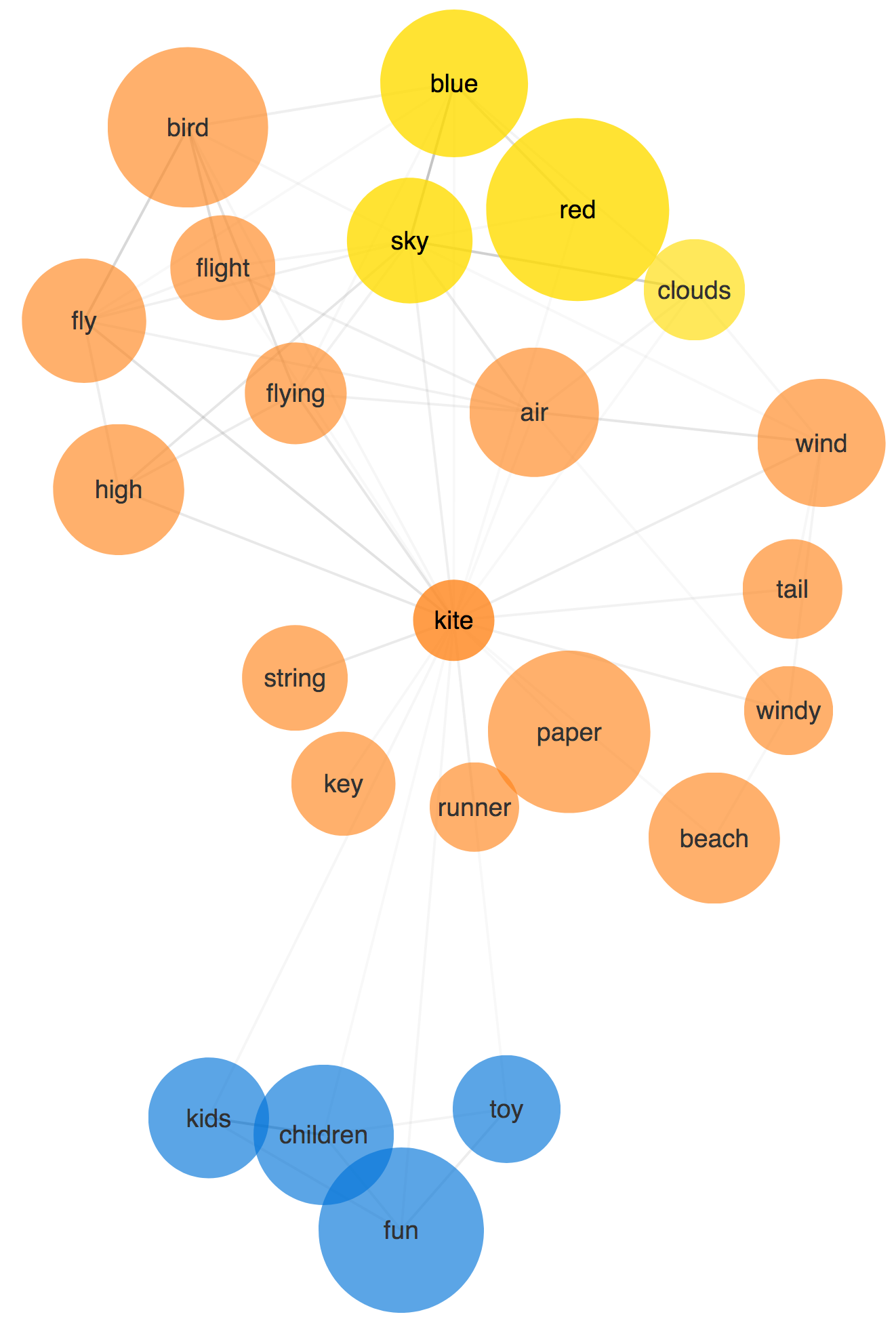}
\vspace{-0.5cm}
\caption{Visualization of the cue {\it kite} and its associated words in the SWOW lexical knowledge base. Source: \url{https://smallworldofwords.org/en/project/explore}}
\label{figure:kite_in_swow}
\end{figure*}

In SWOW each word \textit{w} is mapped into a collection of prominent features as these are named by words elicited from laypersons cued with \textit{w} (example in Figure \ref{figure:kite_in_swow}).
83,863 participants were cued with 12,217 words to respond with 3 associated words, from which 100 responses were collected per cue, resulting in each cue being associated 300 times \cite{de2018small}. We follow the methodology and data in \cite{de2016predicting} to turn this map into a graph, rendered as an adjacency matrix.
In the resulting matrix $A_{G}$, with every word displayed in the rows and columns, a cell $A_{Gij}$ contains the count frequency of word $i$ with word~$j$, the accumulated times that $j$ was responded when $i$ was cued.

\subsection{From lexical graphs to vectors}

To convert graphs into embeddings, we experimented with one outstanding representative from each major family of conversion techniques, viz. based on edge reconstruction, matrix factorisation and random walks. 

\subsubsection{Edge reconstruction}
\label{subcet:edgereconstruction}

Conversion techniques based on edge reconstruction support efficient training but ensure optimisation 
using only local information between nodes that are close to each other in the graph.

They operate on graphs represented as edge lists. An edge is a triple $\langle lhs,rel,rhs \rangle$ where $lhs$ (left-hand side) and $rhs$ are nodes connected by relation $rel$.
The system is trained to recognize triples that are feasible (present in the graph) from the infeasible ones.

As a representative of edge reconstruction techniques, we adopted Semantic Matching Energy (SME) \cite{bordes2014semantic} and used its publicly available implementation.\footnote{
\url{http://github.com/glorotxa/SME}.}

\textbf{Inference-based} With WordNet, the triples were generated this way: for each word $w_{lhs}$ in the vocabulary and for each synset $s_{lhs}$ this word belongs to, a triple is generated for each word $w_{rhs}$ in each synset $s_{rhs}$ (that $w_{rhs}$ belongs to) such that there exists a relation $rel$ between $s_{lhs}$ and $s_{rhs}$, and both $w_{lhs}$ and $w_{rhs}$ are in the vocabulary.\footnote{Data extracted with NLTK \url{www.nltk.org/\_modules/nltk/corpus/reader/wordnet.html}.}

The models were trained for 500 epochs, with evaluation at every 10 epochs, a learning rate of 0.01 and 200 batches. The remaining parameters were the default ones. The validation and test sets each made up for around 5\% of the dataset and the model with the best performance on the validation set was picked.
For a fair comparison among conversion methods, the training data is based on the same 60k vocabulary as in the matrix factorisation in Section \ref{subsect:factorisation} below. The vocabulary was selected with the procedure used in \cite{saedi2018wordnetEmbeddings,brancoEtAl-2019-gwc}, retaining the nodes with the largest number of outgoing edges.

Also for the sake of comparison with the other experiments with cooccurrence-based embeddings in Section \ref{sect:coocurrence}, we chose vectors of dimension 300. 

\textbf{Feature-based} With SWOW, the relations were obtained from the associative strength files that were generated by using the publicly available implementation.\footnote{\url{http://github.com/SimonDeDeyne/SWOWEN-2018}}
The strength file is generated for three association types separately (\textit{R1, R2, R3}), 
which induced three relations that were taken into account as three $rel$ types by the SME method with SWOW \cite{salawa2019thesis}.

We used the same implementation and methodology to obtain SME models as used for WordNet. We empirically chose 
a smaller interval between the evaluations (every 5 epochs instead of 10) and a lower learning 
rate (0.001 instead of 0.01) for better training. 
We took again a vector size of 300.

\subsubsection{Random walk}

Another family of graph embedding methods relies on "text" that results from concatenating the words in the nodes that are visited in a random walk through the graph.
The word embeddings are obtained from some deep learning techniques over that artificial text. 
Starting at a random node, at each iteration, a neighbour node is randomly chosen (with a probability $\alpha$) to be the starting point 
of the next iteration or stopping the walk (with a probability 1-$\alpha$) \cite{goikoetxea2015random}.

Differently from the edge reconstruction and matrix factorisation  approaches, this technique is effective and accommodates global information about the nodes. However, as it only considers 
the local context in a path at each iteration, that makes it hard to stumble on an optimal sampling.

\textbf{Inference-based} 
We used the default Gensim's \cite{rehurek_lrec} Skip-Gram implementation, with a vector dimension of 300.

For the sake of comparability among \KB, we restricted the original technique to use only the edges among nodes 
and to ignore the glosses.
The random walk was applied to the same WordNet graph (60k vocabulary) used with the edge reconstruction and matrix factorization techniques described above in \ref{subcet:edgereconstruction} and below in \ref{subsect:factorisation}.

\textbf{Feature-based} The random walk over SWOW used the same basic setup as used for WordNet.

The SWOW dataset used for edge reconstruction was converted into a graph input for UKB\footnote{ \url{http://github.com/asoroa/ukb/} (default parameters)} and used the default UKB random walk parameters.
To obtain the word embeddings, the default Gensim's Skip-Gram implementation with vectors of dimension 300 was used.

\subsubsection{Matrix factorization}
\label{subsect:factorisation}

A third type of graph embedding method relies on graphs represented by matrices and on their factorization. This is perhaps the family of techniques with the largest number
of instances in the literature, which in many cases result from slight variants
in the tricks used to weight and condense the nodes in the matrix \cite{cai2018comprehensive}.

Matrix factorisation inverts the trade-off found in edge reconstruction. It takes into account the affinity between nodes at the global level of 
the graph, but at the cost of a large time and space consumption.

\textbf{Inference-based} To convert WordNet, we started by building an adjacency matrix with a size above 155k. 
Following \cite{saedi2018wordnetEmbeddings}, we resorted to Katz index for the factorization technique, and used the relevant parameters and other options empirically determined there --- including, for an affordable computational footprint, the same 60k subgraph, made of words with the largest number of outgoing arcs.

After the Katz procedure, 
a Positive Point-wise Mutual Information transformation (PMI+) was applied, to reduce the frequency bias \cite{de2016structure}, followed by L2-norm to normalise each line of the matrix output by the Katz procedure, and finally a Principal Component Analysis (PCA) was applied to reduce the dimension of the vectors, set to 300.

\textbf{Feature-based} The adjacency matrix from SWOW was factorised following the same steps.

Due to the small 12k vocabulary available from SWOW, no extraction of a subset was necessary as it formed a dataset computationally manageable.
 
\subsubsection{PMI}
 
In addition to the matrix factorization method used above, in our experiments we resorted also to a streamlined version of it where the computationally highly costly matrix inversion procedure in the Katz index is skipped, remaining only the PMI transformation, followed by an L2 normalization, and PCA to reduce the matrix size to 300.
 
\textbf{Feature-based} In SWOW, for each pair of cue and associated word, the PMI score was obtained from the number of times they were associated one to the other divided by the product of the number of times each was mentioned normalized by the number of association pairs.

\textbf{Inference-based} In WordNet, for each synset pair related by one edge, the PMI score was obtained as described previously for SWOW, considering a word in a synset as the cue and a word in a synset reached by the edge as an associated word.

\subsection{Coocurrence-based vectors}
\label{sect:coocurrence}

As cooccurrence-based \KB extracted from text, we took the predictive models word2vec \cite{mikolov2013efficient}, GloVe \cite{pennington2014glove}, fastText \cite{bojanowski-etal-2017-enriching}, dependency-based word2vec \cite{levy2014dependency} and the contextual embeddings from BERT \cite{devlin2019bert}.

\section{Brain activity}
\label{sect:brainActivity}

\subsection{Brain activation prediction}
\label{sect:brainActivityMitchell}

The task introduced by Mitchell et al. \shortcite{mitchell2008predicting} consists of predicting the fMRI activation patterns in human subjects from some semantic representation of nouns. To collect the fMRI data used in \cite{mitchell2008predicting}, 9 participants were randomly shown 60 different noun-picture pairs, each presented 6 times.
For each participant, a representative fMRI image for each stimulus was calculated as the mean fMRI response from the 6 repetitions and subtracting the mean of all stimuli.
This task consists of mapping an input lexical semantic representation of each word into an approximation of its vector with the activation values for the 3x3x6~$mm^{3}$  voxels in the fMRI.

\textbf{Training and evaluation} To obtain the prediction models, we resorted to the implementation\ by Abnar et al. \shortcite{abnar2017experiential}.\footnote{\url{https://github.com/samiraabnar/NeuroSemantics}}
The training ran for 1,000 epochs, with a batch size of 29 and a learning rate of 0.001.
The loss function was calculated by adding the Huber loss, the mean pairwise squared error and the L2-norm (on weights and bias).
Like in previous works, only the 500 most stable voxels were selected. 
%The training was done on a Tesla K40m GPU, and took 54 hours (6h per subject) for WordNet embeddings.
We followed the usual evaluation procedure for this task. Separate models were learned for the 9 participants and evaluated using leave-two-out cross-validation, where the model was asked to predict the \fmri activation for the two held-out words in each iteration.
The predictions were matched against the observed activations using cosine similarity over the 500 most stable voxels.

With each word embeddings from Section \ref{sect:commonRepres}, a different model was trained for this task. Evaluation results are in Table \ref{table:evaluationBrainFmri}. Upon empirical experimentation, from BERT (base-uncased), the concatenation of the last four layers provided the best results, and were used in all experiments.

\begin{table*}[h]

    \centering
    \begin{tabular}{lccccc}
        \footnotesize
        \rule{0pt}{13pt}%
         & fastText & word2vec & BERT & GloVe & depend\\
        \hline
        \rule{0pt}{13pt}%
        Cooccurrence-based & 76.57 & 77.78 & 77.98 & 78.59 & \textbf{82.76}\\
        \hline\hline
        \rule{0pt}{13pt}%
        & Edge & Factor. & Walk & PMI\\
        \hline
        \rule{0pt}{13pt}%
        Inference-based (60k) & 61.08 & \textbf{69.42} & 69.37 & 62.72\\
         Inference-based (120k) &  &  & \textbf{72.76} & 72.54\\
         Feature-based & 76.40 & 54.16 & 73.65 & \textbf{81.24}\\
    \end{tabular}
    \caption{\label{table:evaluationBrainFmri}\textbf{Intrinsic evaluation}: Performance with knowledge bases (rows) under different embedding techniques (columns) in terms of accuracy in \textbf{predicting brain activation} with the \textbf{word-rooted fMRI-based data set} (higher is better).
    }
\end{table*}

\textbf{Discussion} The best options with cooccurrence- and feature-based models show a similar level of performance (82.76\% and 81.24\% of accuracy) and are much better than the best option for the inference-based model (72.76\%).

Concerning graph embedding methods, their relative ranking depends on the type of lexical knowledge base to which it is applied. For the inference-based ones (based on WordNet), edge reconstruction is outperformed by matrix factorization, which is on a par or outperformed by random walk. For the feature-based ones (based on SWOW), this ranking appears inverted: the random walk is outperformed by the edge reconstruction --- while matrix factorization appears with an outlier score, way below the scores of any other option whatsoever. Hence,
graph embedding techniques that take into account the affinity between nodes at the global level of the graph are better for inference-based approaches (likely in line with their supporting of transitivity), while for feature-based approaches (not supporting transitivity), techniques taking into account local information between nodes close to each other are better.

Nevertheless, while showing high variability depending on the knowledge base to which it is applied, it is the much more simple PMI (not originally conceived as a graph embedding method but as a mere measure of association) that when applied to the SWOW-based graph, supports the top performance (81.24\%) among all the graph embedding methods.

Focusing on cooccurrence-based models, in turn, these show a good performance in a quite narrow 6 percentage points range, from fastText (76.57\%) to dependency (82.76\%).

All in all the striking empirical finding in this experiment is the very competitive result of SWOW-PMI (81.24\%) --- a dataset only with 12k words, crowd sourced from laypersons cued for the simple associative word retrieval task, and converted into embeddings with the simple PMI technique ---, against the best performing cooccurrence-based KB, dependency word2vec (82.76\%) --- based on a 1.5B corpus, parsed for syntactic dependencies, with a 175k vocabulary and 900k syntactic contexts.

SWOW reflects the frequency of what types of objects in the world are encountered together in typical situations --- rather than the frequency of what words are together in text side by side. Dependency embeddings reflect the frequency of what types of words in the text are encountered together as co-arguments and modifiers in the same predicative structures, which denote typical situations encountered in the world --- rather than the frequency of what words are together in text side by side in linear windows of context.
These considerations seem to indicate that it is this essential aspect of their design, common to both approaches, that lends them their superiority in this task. %On the other hand, 
Given this is a task on brain activity prediction, they seem to indicate also that, because of this, they have a higher cognitive plausibility to represent lexical semantic knowledge than the other options.

\subsection{Further brain activity prediction}

CogniVal is a workbench to test word embeddings on a battery of tasks concerning the prediction of cognitive activity related to language processing \cite{hollenstein2019cognival}. It encompasses 8 tasks for behavioral activity (eye-tracking) and 8 tasks for brain activity (4 with EEG and 4 with fMRI). For a tested word embeddings, CogniVal delivers an aggregated performance score in different tasks.

The 4 fMRI tasks include Mitchell et al.'s \shortcite{mitchell2008predicting} one plus three others, based on the datasets HARRY POTTER \cite{wehbe-etal-2014-aligning}, ALICE \cite{brennan2016abstract} and PEREIRA \cite{pereira2018toward}, which concern the processing of sentences rather than words.

We evaluated all the embeddings used in Section \ref{sect:brainActivityMitchell} above with Mitchell et al.'s \shortcite{mitchell2008predicting} task  also on these 3 sentence-rooted fMRI-based tasks. Results are in Table \ref{table:cognivalFMRIs}.

\noindent\setlength\tabcolsep{3pt}%
\begin{table*}[h]
    \centering
    \begin{tabular}{lccccc}
        \footnotesize
        \rule{0pt}{13pt}%
         & BERT & GloVe  & fastText & word2vec  & depend\\
        \hline
        \rule{0pt}{13pt}%
        Cooccurrence-based & 0.0362 & 0.0256  & 0.0189 & 0.0143 & \textbf{0.0101}\\
        \hline\hline
        \rule{0pt}{13pt}%
        & Edge & Factor. & Walk & PMI\\
        \hline
        \rule{0pt}{13pt}%
        Inference-based (60k) & 0.0117 & 0.0153 & 0.0520 & \textbf{0.0107} \\
        Inference-based (120k) &  &  & 0.0280 & \textbf{0.0113}\\
        Feature-based & 0.0121 & 2.5005 & 0.0288 & \textbf{0.0111}\\
    \end{tabular}
    \caption{\label{table:cognivalFMRIs}\textbf{Intrinsic evaluation}: Performance with knowledge bases (rows) under different embedding techniques (columns) in terms of accuracy in \textbf{predicting brain activation} with the three \textbf{sentence-rooted fMRI-based CogniVal datasets} as mean squared error (\textbf{lower is better}).}
\end{table*}

%\vspace{-16pt}

\textbf{Discussion} A major difference to testing with Mitchell et al.'s \shortcite{mitchell2008predicting} dataset is that, for these 3 sentence-rooted fMRI-based embeddings, while the best cooccurrence-based option is also dependency word2vec --- also supporting now the top performance (0.0101; N.B.:lower scores are better) among all options  ---, the best feature-based option (0.0111, with SWOW-PMI) is closer to the best inference-based option (0.0107, with WordNet60k-PMI) than to the top cooccurrence-based one, and it is even slightly outperformed by this WordNet60k-PMI option.

Another difference is that, while dependency keeps supporting the top performance, the ranking of the other text-based embeddings in the previous experiment (Table~\ref{table:evaluationBrainFmri}) is somewhat inverted: for those coming after dependency, the options that were the first two become now the last two and vice-versa (Table~\ref{table:cognivalFMRIs}).

Yet another difference concerns the relative merits of the graph to embeddings conversion methods. While different techniques were better with different \KB in the previous experiment, now there is one method that outperforms the other two in both \KB, inference- and feature-based, namely edge reconstruction. This method takes into account local information between nodes close to each other.

It is worth noting that while the dataset in \cite{mitchell2008predicting} was collected with each word being processed in isolation by the subjects, in the other three fMRI datasets in CogniVal, the information was collected for words in the context of sentence processing. As from the  experiment with Mitchell et al.'s \shortcite{mitchell2008predicting} task above one noticed that word representations from SWOW and dependency are better at reflecting the frequency of typical situations, this second experiment indicates that SWOW is inferior to dependency at providing such type of cognitively plausible representations when fMRI patterns of words are captured for their occurrence within sentences. Understandably, this should follow from the ways the data in dependency (words in sentences) and SWOW (words in isolation) were collected.

\section{Similarity and relatedness}
\label{section:similarity}

The tasks considered in the third experiment consist of predicting the semantic similarity or relatedness between words in pairs and in seeking 
to match the gold scores assigned by humans to such test pairs --- with the cosine between the predicted vectors mapped into the scale used for gold scores.

The goal in this paper is not to beat the state of the art in similarity prediction tasks but to understand whether when testing embeddings from different empirical sources, these tasks deliver results similar to those delivered by brain activity tasks --- our interest is on gaining insight about whether the similarity tasks, based on simpler and cheaper data, can be used for the same practical purposes of the brain activity tasks, based in much more expensive and hard to get data \cite{salawaEtAl-2019-ranlp}.

For \textbf{semantic similarity}, we used SimLex-999 (with 999 pairs) \cite{hill2016simlex}, WordSim-353-Similarity (203 pairs) \cite{agirre2009study} and RG1965 (65) \cite{rubenstein1965contextual}. 

For \textbf{semantic relatedness},  WordSim-353-Relatedness (252) \cite{agirre2009study}, MEN (3000) \cite{bruni2012distributional} and MTURK-771 (771) \cite{halawi2012large} were used.
The results for WordNet-~, text- and SWOW-based embeddings are in Tables \ref{table:intrinsic_wn}, \ref{table:intrinsic_text} and  \ref{table:intrinsic_swow}, respectively.

\noindent\setlength\tabcolsep{3pt}%
\vspace{-0.5cm}

\begin{table*}[!htbp]
\label{table:intrinsic_wn}
    \centering
    \begin{tabular}{lcccc}
        \footnotesize
        
        & PMI & Edge & Factor. & Walk \\
        \hline
        \textit{Similarity}\\
        \hline
        \rule{0pt}{13pt}%
        Simlex-999 & 28.26 & 39.63$\pm$1.55 & 49.90 & \textbf{50.93}$\pm$0.15  \\
        WordSim-353Sim & 42.61 & 54.93$\pm$2.31 & 50.80 & \textbf{67.40}$\pm$0.30 \\
        RG1965 & 26.38 & 57.70$\pm$4.84 & 57.00 & \textbf{77.50}$\pm$0.95 \\[.2em]
        \hline
        \textit{Relatedness}\\
        \hline
        \rule{0pt}{13pt}%
        WordSim-353Rel & 16.69 & 26.20$\pm$4.10 & \textbf{30.90} & 28.43$\pm$0.76 \\
        MEN & 29.18 & 39.67$\pm$2.55 & 45.00 & \textbf{52.17}$\pm$0.70 \\
        MTurk-771 & 37.22 & 42.40$\pm$1.25 & 52.80 & \textbf{52.90}$\pm$0.50 

    \end{tabular}
    \caption{\label{table:intrinsic_wn}\textbf{Intrinsic evaluation}: Performance with \textbf{WordNet} 60k under graph to embedding conversion techniques (columns) over test sets for \textbf{semantic similarity and relatedness prediction} (rows) in Spearman's correlation coefficient (higher is better), with three runs averaged where relevant. The coverage of the test sets with WordNet 60k: 100\% of Simlex-999; 100\% of WordSim-353 S; 98.0\% of RG1965; 97.6\% of WordSim-353 R; 83.4\% of MEN; 99.9\% of MTurk-771.
    }
\end{table*}
\vspace{-1.0cm}

\noindent\setlength\tabcolsep{3pt}%
\begin{table*}[!htbp]
\label{table:intrinsic_swow}
\label{table:intrinsic_text}
    \centering
    \begin{tabular}{lccccc}
        \footnotesize

        & BERT & GloVe & depend & w2vec & fastText \\
        \hline
        \textit{Similarity}\\
        \hline
        \rule{0pt}{13pt}%
        Simlex-999 & 27.65 & 37.52 & 44.56 & 43.61 & \textbf{49.24} \\
        WordSim-353Sim & 58.66 & 62.98 & 75.88 & 74.08 & \textbf{79.74} \\
        RG1965 & 55.70 & 65.77 & 71.11 & 74.77 & \textbf{81.31} \\[.2em]
        \hline
        \textit{Relatedness}\\
        \hline
        \rule{0pt}{13pt}%
        WordSim-353Rel & 37.72 & 57.09 & 49.23 & 60.97 & \textbf{71.33} \\
        MEN & 51.27 & 61.78 & 67.65 & 69.89 & \textbf{80.87} \\
        MTurk-771 & 44.43 & 63.07 & 62.23 & 65.69 & \textbf{76.13}    

    \end{tabular}
    \caption{\label{table:intrinsic_text}\textbf{Intrinsic evaluation}: Performance of cooccurrence-based word embeddings \textbf{BERT, GloVe, dependency embeddings, word2vec and fastText} (columns) over test sets for \textbf{semantic similarity and relatedness prediction} (rows) in Spearman's correlation coefficient (higher is better).
    }
\end{table*} 
\vspace{-1.0cm}

\noindent\setlength\tabcolsep{3pt}%
\begin{table*}[!htbp]
    \centering
    \begin{tabular}{lcccc}
        \footnotesize

        & Edge & Walk & Factor. & PMI\\
        \hline
        \textit{Similarity}\\
        \hline
        \rule{0pt}{13pt}%
        Simlex-999 & 54.13$\pm$6.20 & \textbf{69.33}$\pm$0.06 & 67.80 &  68.54 \\
        WordSim-353Sim & 77.07$\pm$4.76 & 84.53$\pm$0.06 & \textbf{85.00} & 83.73 \\
        RG1965 & 83.50$\pm$4.50 & 90.23$\pm$0.49 & \textbf{92.90} &  92.48 \\[.2em]
        \hline
        \textit{Relatedness}\\
        \hline
        \rule{0pt}{13pt}%
        WordSim-353Rel & 70.70$\pm$3.68 & 77.73$\pm$0.23 & \textbf{79.30} &  78.50 \\
        MEN & 78.50$\pm$3.90 & 84.27$\pm$0.06 & 87.20 &  \textbf{87.38} \\
        MTurk-771 & 74.77$\pm$4.21 & 81.10$\pm$0.17 & 80.90 &  \textbf{82.39}

    \end{tabular}
    \caption{\label{table:intrinsic_swow}\textbf{Intrinsic evaluation}: Performance with \textbf{SWOW} under graph to embedding conversion techniques (columns) over test sets for \textbf{semantic similarity and relatedness prediction} (rows) in Spearman's correlation coefficient (higher is better), with three runs averaged where relevant.The coverage of the test sets with SWOW: 99.6\% of Simlex-999; 90.6\% of WordSim-353 S; 83.1\% of RG1965; 87.3\% of WordSim-353 R; 89.4\% of MEN; 93.3\% of MTurk-771. 
    }
\end{table*}

\textbf{Discussion}
The major difference to the previous experiments is that now the best option is not obtained with cooccurrence-based embeddings but with feature-based ones, whose best results on the 6 tasks (82.72 average; Table \ref{table:intrinsic_swow}) are much better than the best results with cooccurrence-based ones (73.10 av.; Table \ref{table:intrinsic_text}), which, in turn, are much better than the best results with inference-based ones (55.30 av.; Table \ref{table:intrinsic_wn}).

Another difference concerns the graph to embeddings conversion methods. Now both inference-based and feature-based lexical knowledge repositories are better converted with graph embedding methods that take into account the affinity between nodes at the global level of the graph (matrix factorization or random walk) --- while the combination SWOW and PMI still outperforms all the other options in two of the six tasks, and deliver results very close to the top ones in the other four.

%\vspace{-1cm}

Yet another difference is the ranking among the text-based embeddings. fastText, the third in the %experiment with the
sentence-based fMRI dataset (Table~\ref{table:cognivalFMRIs}), is now first (Table~\ref{table:intrinsic_text}), switching positions with dependency.

This outcome indicates that similarity and relatedness prediction tests are not probing for the same characteristics probed with brain activity prediction tests --- cognitive plausibility ---, and these two classes of tests for embeddings cannot be interchanged with each other.

Additionally, given the quite different rankings of the cooccurrence-based embeddings in the three experiments above, together with their sharp underperformance vis-a-vis feature-based ones, one wonders what the testing them on similarity tasks --- used in the literature to assess their intrinsic quality --- is revealing about them, thus empirically reinforcing the issues raised analytically in \cite{faruqui-etal-2016-problems}.

\section{Extrinsic evaluation}

Results from intrinsic and extrinsic evaluation in NLP are not necessarily aligned with each other as performance increments in intrinsic results may have or not an incremental impact on the intricacies and performance of the larger systems where components happen to be embedded. As intrinsic vs. extrinsic congruence is thus something that has to be determined empirically, it is important to proceed with probing lexical semantics theories with downstream tasks,\footnote{
Examples a.o. of downstream tasks in \cite{rodriguesEtAl-2017-starsem,silveiraEtAl-2012-ieee,costaaEtAl-2012-eacl}.
} for their extrinsic evaluation.
%\subsection{Downstream tasks}
%\label{subsect:tasks}

For multi-task benchmark, we resort to GLUE platform  \cite{wang-etal-2018-glue}, which contains 9 tasks of 3 types, from which we used a subset of 5 tasks with affordable computational footprint: the 2 single-sentence tasks, CoLA and SST-2; 1 of the 3 similarity and paraphrase tasks, namely MRPC; and 2 of the 4 inference tasks, namely RTE and WNLI.

\textbf{SST-2} is a task-based on the Stanford Sentiment Treebank \cite{socher2013recursive} consisting of sentences extracted from movie reviews and human annotation of their sentiment. The model trained on these data has to determine the sentiment expressed in input sentences.
The \textbf{MRPC} task is based on The Microsoft Research Paraphrase Corpus \cite{dolan2005automatically} with sentence pairs automatically extracted from online news and manually annotated as to whether the sentences in each pair are semantically equivalent.
The \textbf{RTE} task resorts to The Recognizing Textual Entailment datasets from online news, which  result  from  annual challenges for the task of textual entailment, and gather the data from RTE1 \cite{dagan2005rte1},  RTE2 \cite{haim2006rte2},  RTE3 \cite{giampiccolo2007rte3}, and RTE5 \cite{bentivogli2009rte5}.  This task consists of predicting whether the premise entails the hypothesis in each test item.
\textbf{WNLI} is a task-based on the Winograd Schema Challenge \cite{Levesque:2012} where  each  example  contains a  sentence  with  a  pronoun  and  a  list  of  admissible  antecedents.   This  task consists of picking  the  right  antecedent.
Finally, the \textbf{CoLA} task is based on the Corpus of Linguistic Acceptability.

%,\footnote{\url{http://nyu-mll.github.io/CoLA}.
%} 
consisting of examples of sentence acceptability judgments taken from books and journal articles on linguistic theory, with each example being a string annotated as to whether it is grammatical.

\subsection{Training and evaluation}
\label{subsect:models}

The goal here is not to reach the top of GLUE leader board. We are interested rather in the extrinsic evaluation of lexical semantics theories, that is in understanding, under comparable circumstances, if and how the embeddings based on different theories have a different impact on these tasks.

To pursue this goal, we adopted an architecture \cite{wang-etal-2019-tell} that accommodates pretrained semantic representations, comprising three levels: the input layer, the shared encoder layers and the task-specific model. 
For the top layer with task-specific information in each downstream task, we used the respective layer from GLUE. This top layer consists of a hidden layer of dimension 512, with the Dropout technique \cite{srivastava2014dropout} with $p=0.2$, and layer normalization \cite{ba2016layer}. The final output layer is a softmax layer.

To obtain the middle layer, encoding sentence semantics, we used a 2-layered biLSTM with dimensionality 1024. Instead of random initialization, the sentence encoder is first trained with one of the best performing pretraining tasks reported in \cite{wang-etal-2019-tell}, namely STS-B.\footnote{
STS-B, The Semantic Textual Similarity Benchmark \cite{CerDALS17}, is a task in GLUE consisting of determining  the  similarity  of  two  sentences  on  a  continuous  scale from  1  to  5.
}

\begin{figure*}[h]
  \centering
  \includegraphics[scale=1.2]{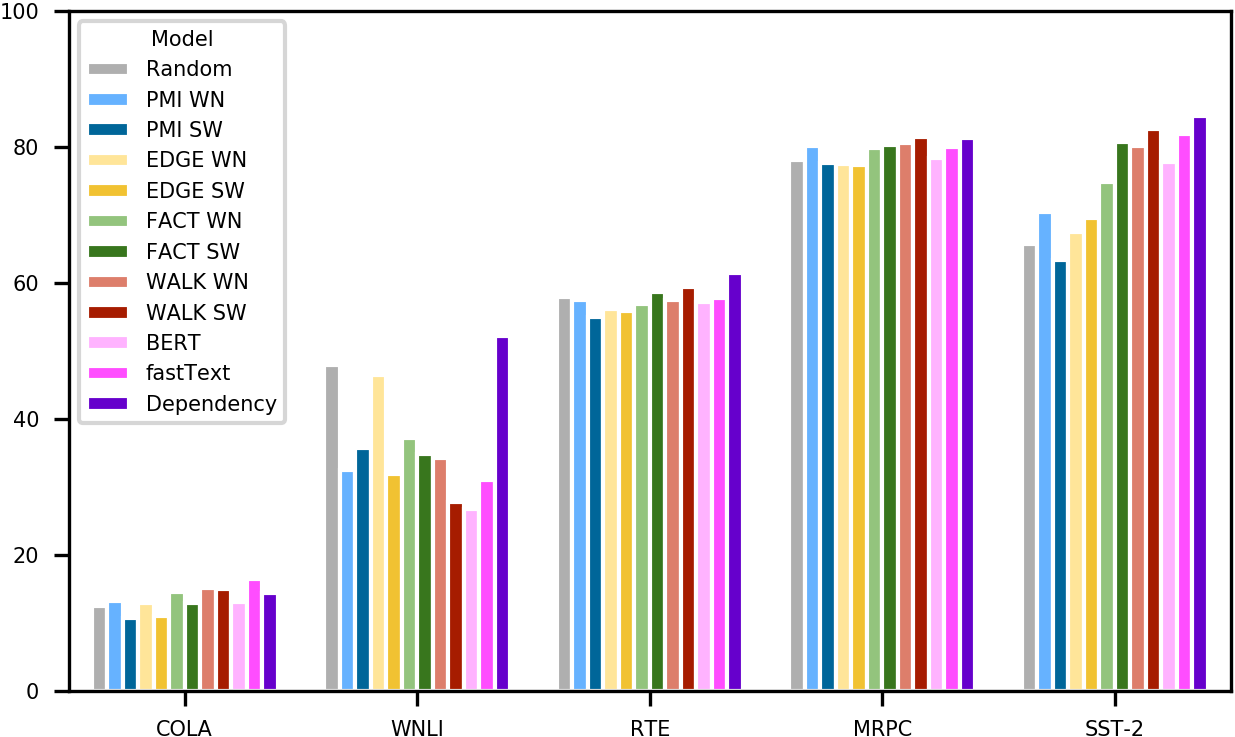}
  \vspace{-0.0cm}
  \caption{\label{figure:extrinsic_intermediate}\textbf{Extrinsic evaluation}: Performance of models with different input layer word embeddings (bars) over the five different GLUE \textbf{downstream tasks} (groups), measured in different evaluation metrics projected to a common range [0, 100]. The reported scores are the average of three runs. The source data for this graph is in Table \ref{table:extrinsic_intermediate}.}
\end{figure*}

For the input layer, we experimented with each of the pretrained word embeddings discussed above in Sections~\ref{sect:brainActivity} and \ref{section:similarity}: for each downstream task, different models were trained with the different pretrained word embeddings, and also with the baseline consisting of embeddings with random vectors. For cross-validation, the framework made use of the original data splitting for training, development and test partitioning.
The results are in Table \ref{table:extrinsic_intermediate} and respective plotting in Figure~\ref{figure:extrinsic_intermediate}.\footnote{We repeated all experiments also with the middle layer contributing sentence semantics removed. The same basic outcome patterns in Figure~\ref{figure:extrinsic_intermediate} were observed again.}

\begin{table*}[htb!]
    % \begin{scriptsize}
    \centering
    \begin{tabular}{llllll}
    \cline{2-6}
    \multicolumn{1}{l}{} &
    \multicolumn{1}{c}{CoLA} & \multicolumn{1}{c}{WNLI} & \multicolumn{1}{c}{RTE} & \multicolumn{1}{c}{MRPC} & \multicolumn{1}{c}{SST-2}  \\
    \hline
    Random      & 12.37 $\pm$ 0.65  & 47.87 $\pm$ 8.80   & 57.90 $\pm$ 2.35 & 78.02 $\pm$ 0.78 & 65.70 $\pm$ 0.95     \\ \hline\hline
    PMI WN      & 13.17 $\pm$ 2.83  & 32.40 $\pm$ 7.58  & 57.40 $\pm$ 0.99  & 80.12 $\pm$ 0.19 & 70.40 $\pm$ 0.59     \\
    PMI SW      & 10.70 $\pm$ 2.35  & 35.70 $\pm$ 8.15  & 54.97 $\pm$ 2.73  & 77.63 $\pm$ 0.47 & 63.30 $\pm$ 1.08     \\ \hline
    Edge WN     & 12.93 $\pm$ 2.11  & 46.47 $\pm$ 13.56 & 56.20 $\pm$ 2.71  & 77.47 $\pm$ 0.80 & 67.47 $\pm$ 0.31     \\ %\hline
    Edge SW     & 10.97 $\pm$ 1.53  & 31.93 $\pm$ 4.28  & 55.83 $\pm$ 1.77  & 77.33 $\pm$ 0.78 & 69.50 $\pm$ 0.90     \\ \hline
    Fact WN     & 14.43 $\pm$ 1.85  & 37.10 $\pm$ 7.74  & 56.93 $\pm$ 1.17  & 79.82 $\pm$ 0.89 & 74.90 $\pm$ 0.52     \\ %\hline
    Fact SW     & 12.93 $\pm$ 2.01  & 34.77 $\pm$ 7.76  & 58.60 $\pm$ 2.71  & 80.23 $\pm$ 1.03 & 80.70 $\pm$ 0.92     \\ \hline
    Walk WN     & 15.13 $\pm$ 1.95  & 34.27 $\pm$ 17.21 & 57.53 $\pm$ 1.81  & 80.62 $\pm$ 0.43 & 80.20 $\pm$ 0.26     \\ %\hline
    Walk SW     & 14.87 $\pm$ 0.15  & 27.70 $\pm$ 3.58  & 59.43 $\pm$ 1.37  & \textbf{81.50} $\pm$ 1.26 & 82.63 $\pm$ 1.12     \\ \hline\hline
    %Glove       & 14.40 $\pm$ 0.78  & \textbf{52.10} $\pm$ 7.27  & \textbf{61.47} $\pm$ 2.56  & 81.30 $\pm$ 0.23 & \textbf{84.57} $\pm$ 0.58     \\
    BERT        & 13.03 $\pm$ 1.97  & 26.77 $\pm$ 3.06  & 57.17 $\pm$ 0.46  & 78.38 $\pm$ 0.24 & 77.80 $\pm$ 0.78     \\
    fastText & \textbf{16.43} $\pm$ 2.25 & 30.97 $\pm$ 6.40 & 57.73 $\pm$ 1.56 & 79.92 $\pm$ 0.48 & 81.87 $\pm$ 0.76 \\
    %Dependency  & 07.80 $\pm$ 2.41  & 31.93 $\pm$ 4.76  & 58.47 $\pm$ 2.32  & 76.50 $\pm$ 0.43 & 71.17 $\pm$ 1.22     \\
    Dependency       & 14.40 $\pm$ 0.78  & \textbf{52.10} $\pm$ 7.27  & \textbf{61.47} $\pm$ 2.56  & 81.30 $\pm$ 0.23 & \textbf{84.57} $\pm$ 0.58     \\
    \hline
    \end{tabular}
    \caption{\label{table:extrinsic_intermediate}\textbf{Extrinsic evaluation}: Performance over five GLUE \textbf{downstream tasks} (columns) of models with different input layer word embeddings (rows). For the task CoLA, performance is measured with Matthews correlation. For MRPC, an average of accuracy and F1 is reported. For the remaining tasks, accuracy is reported. The evaluation scores were projected to a [0-100] common scale (higher is better), with bold denoting top results. Each score is the average of the results from three runs with the random seeds 1147, 1256 and 1179. To enhance the readability of eventual data patterns, the content of this table is rendered in Figure \ref{figure:extrinsic_intermediate}.}
    % \end{scriptsize}
\end{table*}

\textbf{Discussion} Comparing the overall performance across tasks, a first pattern emerges with a group of two tasks, CoLA and WNLI, performing below half the respective absolute best possible score, and the group with the other three tasks performing above.

In that first group, a large majority of the models with pretrained word embeddings perform below or at best on a par with random embeddings. The two tasks in this group are very hard as they rely on rich information about the grammatical structure of the sentences and on long-distance relations among their expressions: to resolve anaphors and find their antecedents somewhere in the sentence, in WNLI; and to categorically decide sentence membership in the language defined by the grammar, in CoLA. Hence, the signal from the lexical information encoded in pretrained embeddings, from whatever lexical theory or empirical source, has an impact that is null, in CoLa, or even detrimental, in WNLI, being of little value to advance the research questions addressed here.

Turning to the other three tasks, the better is the performance of models for a given task, the more they overperform the random baselines.
The top performing datasets are dependency and SWOW-RandomWalk --- on a par in MRPC, and with SWOW-RandomWalk just slightly behind dependency in RTE and SST-2. 

Concerning feature-based embeddings, these results confirm, also in extrinsic evaluation tasks, their strength found in intrinsic tasks --- where SWOW supported the best performing options in semantic similarity and in word-rooted brain activity prediction tests. 
Given the consistent superiority of SWOW embeddings in intrinsic tasks, and their very competitive performance here in dowstream tasks, these results indicate that SWOW-based embeddings are likely one of the most reliable candidates to be used to enhance NLP downstream tasks for which the contribution of (pre-trained) lexical knowledge is useful.

Concerning text-based embeddings, dependency (top in brain activation prediction) outperforms fastText (top in similarity), thus in line with their relative ranking in the brain task, but inverting that relative ranking in the similarity task. This reinforces the wondering about how useful NLP intrinsic tasks may be to assess the text-based embeddings for their strength in downstream tasks, in line with \cite{chiu-etal-2016-intrinsic}.

Concerning graph embedding methods, taking PMI aside, here edge reconstruction is overperformed by matrix  factorization, which is overperformed by random walk. Methods taking into account the affinity between nodes
at the global level of the graph seem thus better for extrinsic tasks.
Another interesting lesson is that, for each graph embedding method, models using SWOW embeddings overperform models using WordNet embeddings --- except for PMI, better now with WordNet than with SWOW.

\section{Conclusions}

\textbf{Contributions} A first major contribution of this paper is \textit{the design of an experimental setup to comparatively probe lexical semantics theories} of all kinds --- feature-, inference- and cooccurrence-based --- for both their cognitive plausibility and their usefulness in NLP. 

This setup consists of converting representative repositories of these theories to a common format and integrating them, under such a common format, in different models for different probing tasks: \textbullet~To be converted, exemplar techniques from each major graph embedding family (edge reconstruction, matrix factorization, random walk) were used. \textbullet~To be probed for cognitive plausibility, some of these models addressed fMRI-based brain activation prediction tasks. \textbullet~To be probed for usefulness in NLP, some other models addressed semantic similarity and relatedness prediction tasks, for intrinsic evaluation, and they were also embedded in downstream NLP tasks, for extrinsic evaluation.

Another major contribution is the empirical results obtained with a systematic application of this probing setup, including the central finding that \textit{the feature-based lexical knowledge base is superior to knowledge bases complying with other lexical semantic theories} in consistently supporting models with top performance across the different probing tasks. 

Concerning the other, incidental empirical findings: \textbullet~As to graph embedding techniques, it was possible to understand 
%\textit{which embedding techniques are better suited for which tasks and resolvers}, as identified in the paper.
that \textit{graph embedding methods taking into account the affinity between nodes at the global level of the graph tend to better serve downstream tasks}. \textbullet~As to intrinsic vs. extrinsic NLP evaluation tasks, there emerged not an alignment between a superior performance in the former and a superior performance in the latter, hence \textit{intrinsic performance of a lexical knowledge base is an unreliable predictor of its extrinsic performance}. \textbullet~As to text-based embeddings, top performance models in different tasks are supported by different embedding methods, hence \textit{the  performance of a given method for text-based embeddings in one task is an unreliable predictor of its performance in other tasks}.

\textbf{Discussion} The overall superior performance of the feature-based SWOW lexical knowledge base is the striking empirical result of the present study. This knowledge repository supported the top results, without a close second, in semantic similarity and relatedness prediction. It was on a par to top results or a close second in extrinsic tasks where the lexical signal has an impact. It was a close second in word-rooted brain activity prediction. As expected, it was not shining only in sentence-rooted brain activity prediction.

It only adds to its outstanding record that it is uncertain how comparability can be fairly ensured between SWOW and its alternatives. Dependency embeddings are extracted from a 1.5B corpus, with a 175k vocabulary (word2vec$/$100B, fastText$/$600B, GloVe$/$840B).
SWOW has 12k words linked to each other whose links were crowdsourced from 83k laypersons cued 3 times for basic word association.

It also adds to the overwhelming performance of SWOW that it comprises just 12k words while the lexicon of an adult is estimated to consist of over 40k words \cite{rysbaert2016mentalLexicon40k} --- which additionally emphasizes its yet untapped potential and the promise of extended versions with larger numbers of words.

To explain its strength, it is tempting to attribute it to SWOW's higher cognitive plausibility. This bears the underlying assumptions --- epistemologically non trivial --- that better performance at the word-rooted brain activity prediction correlates with higher cognitive plausibility of lexical semantic knowledge repositories, and higher cognitive plausibility correlates with better performance of NLP systems where these repositories are embedded.

\textbf{Future work} Returning to one of our driving questions: could there be a unified account of lexical semantics such that the three families of approaches to lexical meaning seamlessly emerge as (partial) renderings of (different) aspects of the same core semantic knowledge base?
The findings reported in this paper make a positive answer to it increasingly  attractive, with feature-based lexical knowledge being a good candidate to that core knowledge repository.

It is worth noting though that one has a grasp on how to obtain cooccurrence-based lexicons from feature-based ones 
(De Deyne et al., 2016b
%\cite{de2016predicting} 
and the present paper); and also
from inference-based ones 
(Saedi et al., 2018 and the present paper)
%\shortcite{saedi2018wordnetEmbeddings}
and vice-versa \cite{Tarouti2016embbedings2wordnet}.
But whether it is possible to go from text-embeddings like fastText to feature-based lexicons like SWOW, or from SWOW to ontologies like WordNet remain open questions. 
Seeking to address these challenges is future work that should help to further progress towards finding an answer to the research question above.

\section*{Acknowledgements}

The research reported here was partially supported by PORTULAN CLARIN---Research Infrastructure for the Science and Technology of Language, funded by Lisboa 2020, Alentejo 2020 and FCT---Fundação para a Ciência e Tecnologia under the grant PINFRA/22117/2016.

\section*{Reproduction}

To support the reproduction of research results \cite{branco2017lre_reproduction}, the embeddings are available from \url{https://hdl.handle.net/21.11129/0000-000D-C67B-A }

\bibliographystyle{coling}
\bibliography{anthology,acl2020}

\clearpage

% \vspace{-0.8cm}

% \vspace{-1cm}

\end{document}